# Automated Wheat Disease Detection using a ROS-based Autonomous Guided UAV

**Behzad Safarijalal**[1,+], **Yousef Alborzi**[1,2,+], **and Esmaeil Najafi**[1,*]

[1]K. N. Toosi University of Technology, Mechanical Engineering , Tehran, Iran
[2]University of Manitoba, Mechanical Engineering, Winnipeg, Canada
[*]najafi.e@kntu.ac.ir
[+]these authors contributed equally to this work

## ABSTRACT

With the increase in world population, food resources have to be modified to be more productive, resistive, and reliable. Wheat is one of the most important food resources in the world, mainly because of the variety of wheat-based products. Wheat crops are threatened by three main types of diseases which cause large amounts of annual damage in crop yield. These diseases can be eliminated by using pesticides at the right time. While the task of manually spraying pesticides is burdensome and expensive, agricultural robotics can aid farmers by increasing the speed and decreasing the amount of chemicals. In this work, a smart autonomous system has been implemented on an unmanned aerial vehicle to automate the task of monitoring wheat fields. First, an image-based deep learning approach is used to detect and classify disease-infected wheat plants. To find the most optimal method, different approaches have been studied. Because of the lack of a public wheat-disease dataset, a custom dataset has been created and labeled. Second, an efficient mapping and navigation system is presented using a simulation in the robot operating system and Gazebo environments. A 2D simultaneous localization and mapping algorithm is used for mapping the workspace autonomously with the help of a frontier-based exploration method.

## Introduction

As the world's population increases, the demand for nutrition resources increases, while in many places, there are no productive farmlands left to use. As a result, the need for ways to produce more on established farmland and more products from fewer resources is one of the present concerns of mankind. One solution is to utilize more efficient technologies in agriculture and use robots instead of human labor in specific tasks to increase production. These include crop scouting, pest and weed control, harvesting, targeted spraying, and sorting. These robots can be classified into two main categories in general: unmanned ground vehicles and unmanned aerial vehicles.

Like many other new technologies, the first use of a unmanned aerial vehicle (UAV) was for military purposes[1]. However, today the applications of these aerial robots can be found in many civilian sectors, such as agriculture. They are used in almost any kind of agricultural tasks such as crop monitoring and disease detection[2,3], crop height estimation[4], pesticide spraying[5] and soil and field analysis[6]. Although recent developments in robotic research have made these machines very smart and effective, the need for further improvements to make them more efficient and more intelligent is obvious and unavoidable.

Regardless of the plant's type, pests have always been one of the biggest concerns of farmers. The attempts for increasing final products by eliminating destructive pests or weeds are age-old concerns. Recent developments in chemical pesticides (mainly after world war II), have brought food safety to humans by removing these irritating creatures and decreasing food loss. On the other hand, the overuse of chemical substances and especially pesticides, has raised concerns about their consequences on human health, which have caused global efforts for chemical usage reduction in food industries. One of the major outcomes of these efforts is a process called integrated pest management (IPM). IPM is concerned with controlling and reducing the number of pests to an acceptable extent, by performing effective, economical, and feasible methods and actions. Some of these actions include using resistant varieties, and natural predators and parasites instead of chemical substances[7]. However, sometimes the employment of pesticides is inevitable. In these cases, using a precise and smart system, which can minimize the amount of used chemical pesticides, by on-time plant disease detection and accurate pesticide spraying on the infected plants, instead of spraying the whole farmland, can help produce more healthy products.

Rusts are one of the most common wheat diseases in the world. In general, three types of rust diseases can damage wheat crops: stem rust, stripe rust, and leaf rust. Wheat leaf rust, caused by Puccinia Triticina, leads to a loss in product yield and quality. Also known as brown rust, this disease is usually less harmful than the other types of rust. However, because of its widespread existence, it amounts to more annual damage overall[8]. In some regions, the amount of damage caused by this type of disease can reach values of up to 70%[9]. Stripe rust, also known as yellow rust, is caused by Puccinia Striiformis. Regional yield losses due to stripe rust, have been reported to reach amounts of 25% while this amount rose to numbers as high as 80% during the Middle East and North African epidemic in 2010[10]. While less common, stem rust or black rust, caused by Puccinia Graminis, can result in serious losses up to 33% of the final products[9]. The amount of inflicted damage by these rusts mainly depends on species resistance and climate situations. The purpose of this work is to propose a smart UAV system, which can monitor a wheat farmland autonomously. This system consists of two parts: an autonomous control and navigation system, and a wheat rust detection system, to monitor the wheat farmland and detect the corrupted crops respectively. The control and navigation system is implemented and tested in a simulated environment using the robot operating system (ROS) and Gazebo. A preliminary version of this work has been already published in[11]. The paper is organized as follows. First, An image-based wheat disease detection and classification system is presented, followed by a description of the implemented quadrotor system structure and algorithms. Finally, the performance of the quadrotor system and the results of the proposed deep learning image classification methods are presented and discussed.

## Methods

The on-time detection of plant diseases is of vital importance in decreasing losses in crop yield. Although laboratory testing is the most accurate way of detecting diseases in plants, it is not always available to farmers and costs a substantial amount. However, because most plant diseases are associated with some visible feature, studying these visual patterns on the plants is a less expensive way of detecting the infections.

### Plant disease detection
Recent advances in image processing and deep learning algorithms have inspired their popularity and drove many teams and researchers to use these techniques for various applications. Additionally, the rise of deep learning algorithms has caused improvements in the accuracy and speed of many problems, especially in image classification. This has led to a great number of researches for Automatic plant disease detection systems based on image classifiers. While earlier works such as[12] use conventional image processing techniques to extract features such as color, important edges, and histograms to detect plant diseases, machine learning algorithms utilize available data to implement more robust models. For example, in[13], K-means clustering along with feature extraction and an SVM classifier have been used to detect diseases in grape images with an 88.89% accuracy.

More recently, with the help of newly developed Graphical Processing Units (GPU), deep learning techniques, especially CNNs, have experienced great success in image classification problems[14] and have become the state of the art in computer vision. Their success on plant disease detection datasets is shown in several previous works. In[15], five different CNN models have been trained and tested on a large dataset, called Plant Village, with 58 classes of plants and diseases. The authors of[16] used variations of the MobileNet[17] architecture to achieve excellent results on the Plant Village dataset. In[18] the VGG-16 architecture has been used to identify five major diseases in eggplant leaves based on a novel dataset. The authors also analyzed the effects of using images in different colors spaces on the accuracy of the system. A different CNN-based approach has been implemented on a custom dataset of different plants in[19], in which individual lesions and spots on the leaf are considered for disease detection, rather than using the whole leaf as input. Furthermore, hyperspectral imaging has also proven to be effective for plant disease detection in different scales in papers such as[20] and[21]. This work proposes a two-stage classifier to enable our UAV to detect plant diseases. First, the model uses an object detection network to find individual plant leaves in the image, and then after cropping the image with the bounding box coordinates, the second classifier detects the type of the disease on the leaf.

*Dataset description*
To the best of our knowledge, there is no open-source dataset of wheat plants with the mentioned diseases available. Therefore, to train our models, we created a custom dataset. The dataset contains 900 images of healthy and infected wheat plant leaves captured in real cultivation conditions in commercial farms. The images were first annotated by hand by bounding boxes for the object detection algorithms. In each image we have annotated as many leaves as possible by specifying their region and labeling all of them under one class regardless of their visual health condition. And for the second-stage classifier, 3672 images of individual leaves were cropped by hand and labelled with their corresponding disease for each class. Finally, they were split into train, validation and test sets randomly. A sample from every class is shown in Fig. 1 and the dataset is described in more details in Table 1.

*Wheat leaf detection*
To detect individual leaves in the image, we require an object detection network. Currently, the state of the art in object detection networks are the YOLO V4[22] and EfficientDet[23] models. The YOLO V4 model consists of a CSPdarknet53[24] backbone and the YOLO V3[25] head for the detector while also implementing different Bag of Freebies and Bag of Special methods such as Mosaic data augmentation, DropBlock[26] and CutMix[27] regularization. In the Efficientdet paper, by utilizing the Efficientnet architecture[28] as the backbone and a bi-directional feature pyramid network (BIFPN) as the feature network and prediction networks, the authors were able to achieve similar accuracies as state-of-the-art models while being 4x-9x smaller. Here we have implemented 4 different object detection networks and trained them on our annotated dataset. Because the features of our plant diseases are relatively small, the network requires images with somewhat high resolutions, therefore, we have trained the networks with $800 \times 800$ resolution images. the training parameters are kept constant for all of the models so that the results can be compared fairly.

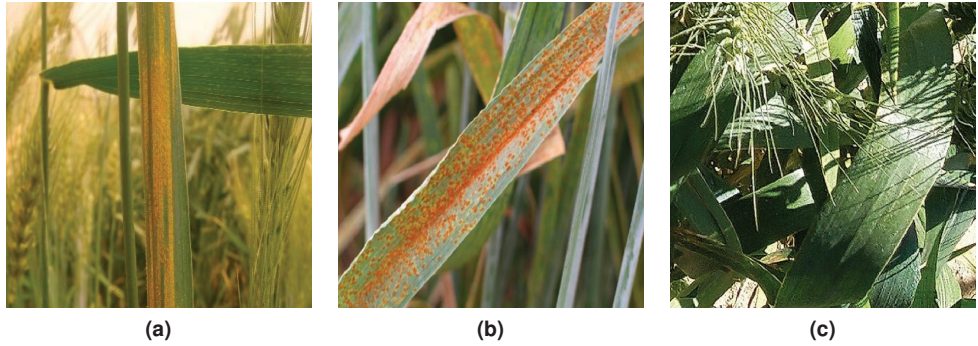

**Figure 1.** Samples from the collected wheat disease dataset: (a) yellow-rust-infected wheat leaf, (b) brown-rust-infected wheat leaf, (c) healthy wheat leaf

| Set | Brown Rust | Yellow Rust | Healthy |
|---|---|---|---|
| Train | 894 | 928 | 1116 |
| Valid | 111 | 116 | 139 |
| Test | 112 | 116 | 140 |
| Total | 1117 | 1160 | 1395 |

**Table 1.** The details of each class and set of our wheat disease dataset

*Wheat rust classification*
After the leaf bounding boxes are detected on the UAV acquired image, they are cropped and fed to the classification network. For the classification part of the detection system, we tested multiple approaches to get the best results while keeping in mind the computation cost. With their weight-sharing ability, CNNs are the best candidate for image classification. In this work, we implemented five different CNNs which were MobileNet V3, VGG 16, Inception, ResNet 50, and EfficientNet-B0. The models are trained for 50 epochs using the adam optimizer with an initial learning rate of 0.001 and on each epoch, the learning rate is divided by a factor equal to the epoch count. Furthermore, data augmentation techniques were used to increase the number of training data by randomly applying rotation, zoom, brightness change, and shearing.

## UAV navigation and control
Autonomous navigation of a robot in an arbitrary location requires various tools and consists of different processes. These components can vary in different cases based on many parameters such as the type of the robot, its workspace (indoor or outdoor), and utilized sensors. Probably the first requirement is a map, which contains the static obstacles of the robot's workspace. This map can be a 2D or 3D representation of the robot's working environment and is acquired in a process called mapping. It is also possible to obtain this map using the robot itself with the help of another process called localization, which enables the robot to map its environment while simultaneously estimating its own position regarding this map. This estimation gets more accurate as the robot moves around and gathers more information about its surrounding. The tasks of mapping and Localization are performed using sensors such as a camera, a laser range-finder, or an ultrasonic sensor, which can help the robot evaluate its distance to the nearby obstacles. By executing mapping and localization together at the same time, the robot is able to map its working environment. This simultaneous performance is referred to simultaneous localization and mapping (SLAM)[29]. Furthermore, this job can be done by manually moving the robot through its workspace until the acquired map is completed, or it can also be done autonomously. Autonomous SLAM is possible with the help of Exploration, in which the robot explores its environment automatically, to map the whole area.

All of the mentioned techniques can be implemented and tested in a simulation environment. The robot operating system ROS with the help of Gazebo and Rviz can be utilized to do such a simulation. ROS is an open-source operating system with several styles of communication and great package management. In addition to lots of packages for almost any type of robotic applications, ROS supports many programming languages such as C++ and python. The main purpose of ROS's development is to support code reuse in robotics research, so anyone could be able to use and improve its materials. This has caused it to become one of the most popular robots simulating environments and its community is growing every day. Furthermore, Gazebo is an excellent free tool that can be used to create various indoor and outdoor environments (worlds) and visually simulate robot movement and navigation in these environments.

*Localization and mapping*
The employed drone for this work must be able to carry the payload of an appropriate RGB camera and a laser range-finder to perform the detection and navigation tasks respectively. In this case, a simple and suitable choice would be Parrot A.R. Drone 2.0. There are many packages available regarding robot simulation for ROS. The tum-simulator package provides a good simulated model of Parrot A.R. Drone 2.0. In this study, this model is used along with hector-quadrotor package, to simulate the drone and implement the requisite systems for autonomous navigation. Hector-quadrotor, provided by Team Hector, is one of the most popular and powerful packages regarding the simulation and navigation of drones[30]. Moreover, reducing the computation costs of the navigation and control process as much as possible is a crucial task. To achieve this goal, using 2D SLAM and navigation methods instead of 3D ones, could be a more efficient and sufficiently accurate approach. So, the "gmapping" and "amcl" packages were utilized to implement OpenSlam's Gmapping and Monte-Carlo localization[31] algorithms in ROS respectively. The "gmapping" package can construct an occupancy grid map efficiently using data provided by the mounted laser rangefinder sensor, based on Rao-Blackwellized particle filters[32]. Also, the "amcl" package can perform the localization task by the adaptive (or KLD-sampling) Monte-Carlo localization method.

*Navigation*
There are two path planners required for the navigation phase, namely global and local planners. The adapted move-base package uses the A* algorithm[33] to compute efficient paths toward goal points, which can handle static obstacles specified in the constructed map. Moreover, the "dwa-local-planner" package handles the dynamic obstacles, which are not provided in the map, by computing appropriate velocity commands using the dynamic window approach[34]. Finally, by using the "exploration-lite" package[35] along with the earlier mentioned packages, the mapping process can be done autonomously. This package's functionality is based on frontiers[36], which are specific points in the uncompleted occupancy grid maps that are between unoccupied and unmapped cells. The robot tries to move toward these frontiers to map the unknown environments, until there are no more frontiers, and thus no more unmapped areas left in its workspace. As mentioned before, the utilized packages for the robot navigation in this work are all 2D. To be more specific, they are designed for UGV navigation, and thus unable to control the drone's height. To control the robot's altitude, a node was written to lift the drone by sending appropriate velocity commands and keep the drone's altitude fixed during its flight using a PID controller.

*System architecture*
In this ROS-based system, different processes are carried out within ROS nodes, which interact with each other using ROS topics. As presented in Fig. 2, the autonomous mapping procedure is done by the cooperation of the "slam-gmapping" and "explore" nodes. By acquiring the drone's current pose via the "tf" topic and the laser rangefinder measurements via the "scan" topic, the "slam-gmapping" node updates the partially-constructed map. On the other hand, by subscribing to the "tf" and "map" topics, the "explore" node, finds the frontiers and sends pose goals to the "move-base" node. The "move-base" node computes the appropriate velocity commands for the robot and creates the local and global costmaps. Finally, the velocity commands are sent to the gazebo node for visualization purposes.

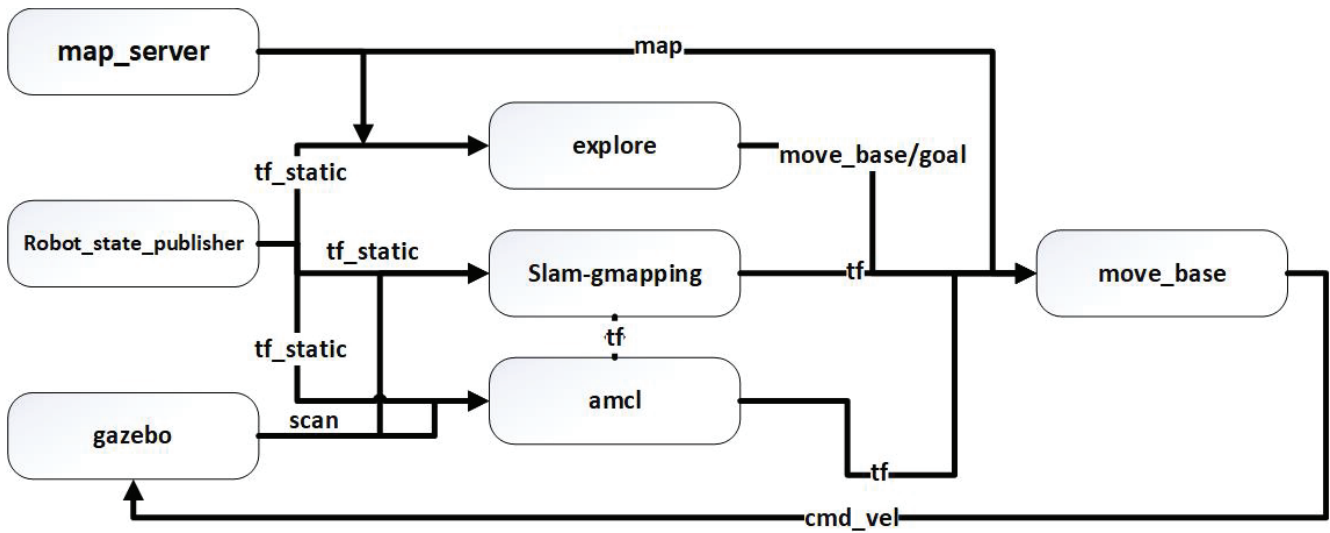

**Figure 2.** UAV navigation system architecture in ROS

| Model | mAP | FPS | BFLOPs |
|---|---|---|---|
| YOLOV4 | 15 | 13.3 | 220.278 |
| YOLOV4 tiny | 19 | 42.3 | 25.101 |
| YOLOV4 tiny-3 yolo layers | 16.50 | 41.6 | 29.652 |
| Efficientdet-B0 | 10.65 | 16.4 | 13.569 |

**Table 2.** Results of different object detection models on our wheat disease dataset

## Results

With the integration of the disease detection system and efficient UAV navigation and mapping, farmers can autonomously monitor their lands for diseases. In this process first, a 2D map of the land is acquired which contains the location and type of the infected crops. This information can then be used to perform appropriate actions, such as spraying a pesticide. Although we designed and trained our detection networks on a dataset for wheat diseases, the proposed system can easily be adapted to any other kind of plant by simply using the appropriate dataset. The results of the proposed wheat disease detection and UAV navigation systems are presented in this section.

### Image-based disease detection

In Table 2, we report the metrics of the different implemented object detection networks. Taking into consideration the importance of speed for our system, we conclude that the YOLOv4 tiny model is the best fit. It is clear that the YOLO V4-tiny model achieves the best results with the highest mAP and FPS. An instance of this model on a sample of our dataset is displayed in Fig. 3.

Next, Table 3 compares the different CNN models in terms of performance and their accuracy on the disease classification test set. Looking at Table 3, we can see that, while the MobileNet-V3 model requires fewer floating point operations for running an instance, the EfficientNet-B0 model has the best combination of accuracy and GFLOPs among all of the networks. For this model, we present the evaluation matrix containing the precision, recall and F1-score, in Table 4 and the confusion matrix in Fig. 4. The confusion matrix shows that the EfficientNet-B0 model only had a single incorrect prediction in the whole test set.

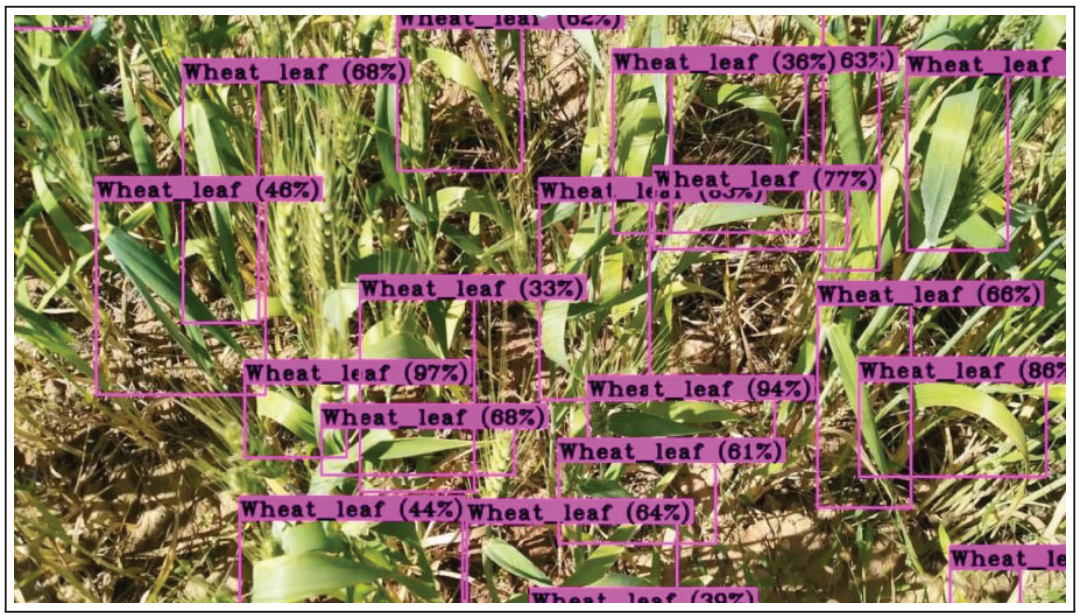

**Figure 3.** Instance of leaf detection from YOLOV4-tiny model

| Model | Accuracy | GFLOPs |
|---|---|---|
| MobileNet V3 | 94.29 | 0.174 |
| Inception | 98.64 | 5.69 |
| ResNet-50 | 98.91 | 7.75 |
| EfficientNet-B0 | 99.72 | 0.794 |
| VGG16 | 99.45 | 31 |

**Table 3.** Results of different CNN classification models on our wheat disease dataset

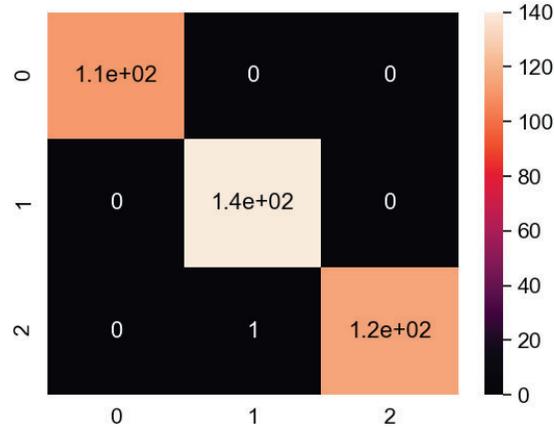

**Figure 4.** Confusion matrix for EfficientNet-B0 model on our wheat disease dataset

| | Precision | Recall | F1-Score |
|---|---|---|---|
| Brown-rust | 1.00 | 0.99 | 1.00 |
| Yellow-rust | 0.99 | 1.00 | 1.00 |
| Healthy | 1.00 | 1.00 | 1.00 |
| Accuracy | | | 1.00 |

**Table 4.** EfficientNet-B0 Matrix of Evaluation

## Quadrotor mapping and navigation

To evaluate the mapping and navigation system, we carried out tests in the "willow-garage" gazebo simulated world. A visualization of the localization process using the "amcl" package is shown in Fig. 5 and as evident by the particle cloud condensation displayed in Fig. 5, after some movement by the robot, the uncertainty of the robot's pose decreases and the localization estimation becomes more accurate. Moreover, with the combination of the generated frontiers by the "exploration-lite" package, an accurate localization, and range measurements from the laser sensor, the SLAM algorithm acquired a map of the simulated environment as shown in Fig. 6. While the simulation environment appears much more complex than a wheat farmland, the mapping system was able to generate the map autonomously without any interference using the exploration package.

Using the constructed map and the implemented localization system, the UAV is able to readily navigate between two arbitrary points. Fig. 7 shows a successful instance of the A* path planning algorithm for an arbitrary goal pose in the simulated environment. Permission to reuse figures 5, 6 and 7 from the preliminary version of our work in[11] has been obtained from the publisher under the IEEE license number 5246641444156.

## Discussion

For the object detection part of the system, by looking at Table 2, we can see that the mAP values of the trained networks seem low compared to the results of these networks on other datsets. This is mostly due to the fact that the desired objects in our dataset are close in color and shape to their background and it generally is a difficult task to detect the individual leaves in the scene. Also, in each image in the dataset there are a large number of wheat leaves present and there is the possibility of missing a few leaves in the labeling process which would hurt the mAP score of the object detection networks. Nevertheless, for our application we do not need to detect every single leaf in the image captured by the UAV. Therefore the networks seem to perform sufficiently in detecting enough leaf samples in an image, as shown in Fig. 3.

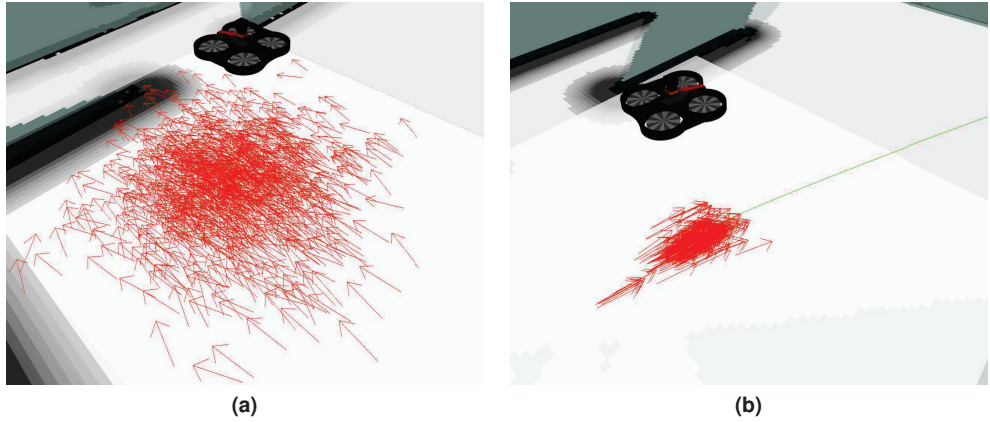

**Figure 5.** Visualized particle clouds condensation process of the "amcl" package (a) initial state (b) state after movement

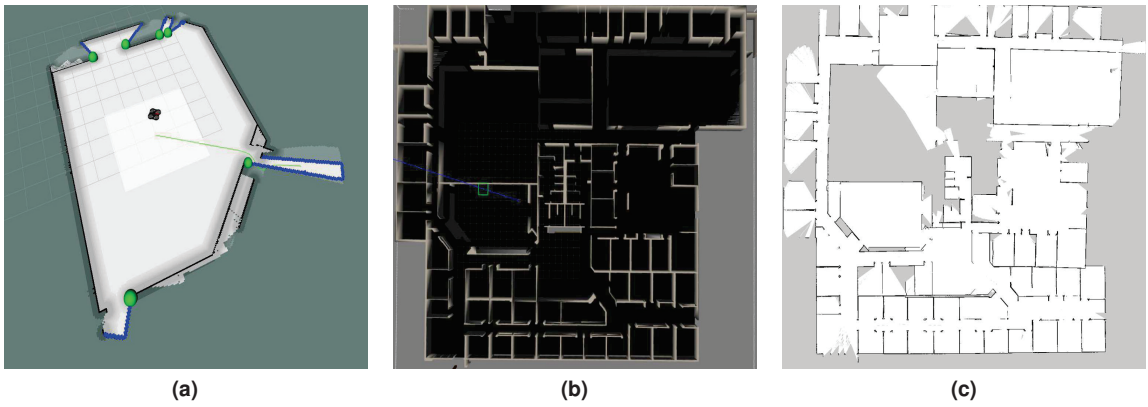

**Figure 6.** Exploration and mapping process : (a) Frontier points detected by the implemented algorithm (green points), (b) Simulated environment in Gazebo, (c) constructed map with frontier exploration

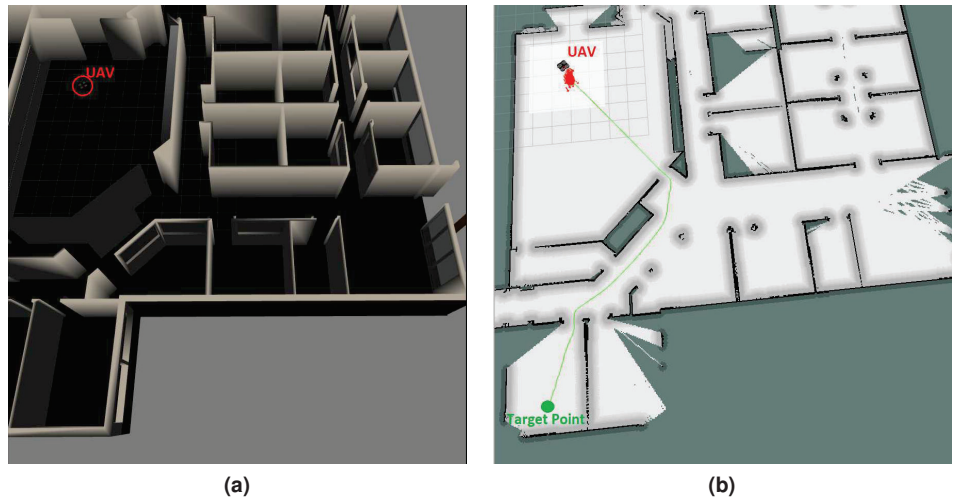

**Figure 7.** Simulated UAV navigation result: (a) UAV in simulated Gazebo world (b) arbitrary navigation path (green contour) in the same simulated environment

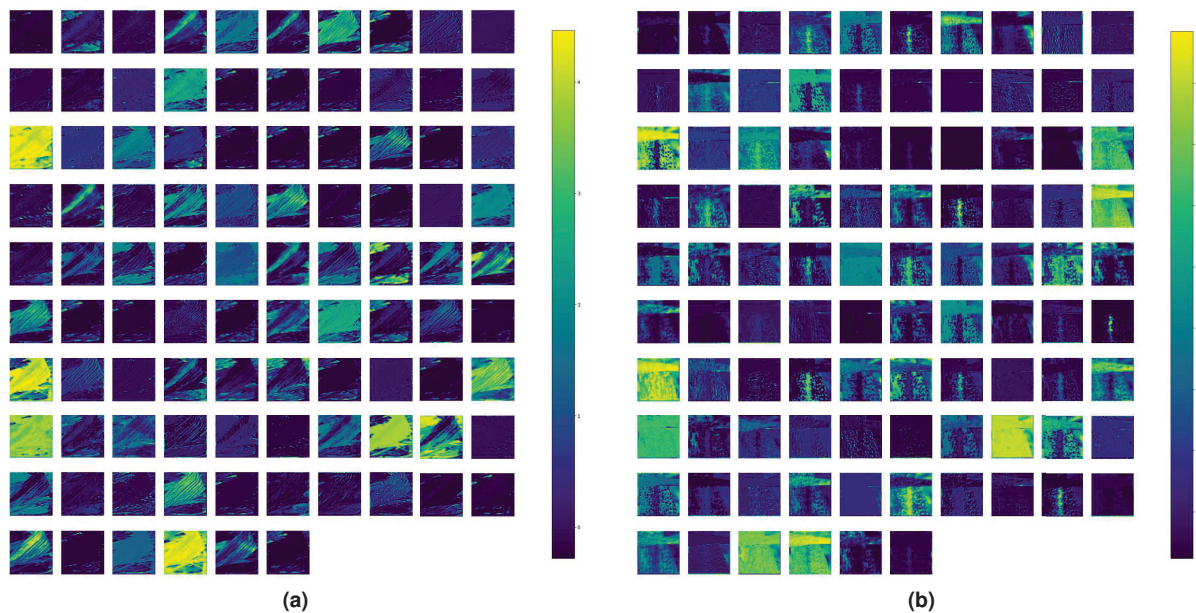

**Figure 8.** Feature map of the EfficientNet-B0 CNN model of (a) yellow-rust-infected leaf , (b) brown-rust-infected leaf

Table 3 shows that the classification networks achieve high accuracies on the test set after the 50 epochs of training. Additionally, visualizing the intermediate feature maps of different blocks of the model can shed light to what the network is actually learning. In Fig. 8, the feature maps of the second convolutional block of the efficientNet-B0 model are displayed. These figures show that the network has been able to learn to detect dotted features for brown rust infected leaves and striped shape features in Yellow-rust infected leaves.

## Conclusion

In this work, a smart wheat crop monitoring system has been proposed. The image-based classification system can detect individual leaves in the images captured by the UAV mounted camera and classify them into three groups of brown-rust-infected, yellow-rust-infected, and healthy leaves. After evaluating different object detection algorithms on our custom dataset, the YOLOV4-tiny model outperformed other algorithms and achieved the best results with a 19% mAP 0.5 and 42 frames per second frame rate. For the disease classification system, the performance of the EfficientNet-B0 model was superior to other tested CNN architectures, recording the highest accuracy of 99.72% and 0.794 GigaFLOPs. Finally, an efficient navigation system based on the A* algorithm was implemented on the UAV, which can find safe trajectories between two arbitrary points in the map. The required map of the environment is also acquired autonomously by using the SLAM algorithm and the frontier exploration technique. This navigation system was simulated in Gazebo using the robot operating system.

## Data Availability

The dataset generated and analysed during the current study are available in the Kaggle repository via the following web link: https://www.kaggle.com/sinadunk23/behzad-safari-jalal